\documentclass[conference]{IEEEtran}
\usepackage{cite}
\usepackage{amsmath,amssymb,amsfonts}
\usepackage{algorithmic}
\usepackage{graphicx}
\usepackage{textcomp}
\usepackage{xcolor}
\def\BibTeX{{\rm B\kern-.05em{\sc i\kern-.025em b}\kern-.08em
    T\kern-.1667em\lower.7ex\hbox{E}\kern-.125emX}}
    
\usepackage{bbm}
\usepackage{bm}
\usepackage{mathtools}
\usepackage{autonum}
\usepackage{tabularx}
\usepackage{enumitem}% http://ctan.org/pkg/enumitem
\usepackage{subfig}
\usepackage{caption}
\usepackage[numbers]{natbib}
\usepackage[flushleft]{threeparttable}
\graphicspath{{images/}}
\newcommand*{\aucroc}{\operatornamewithlimits{AUC-ROC}}
\newcommand*{\ndcg}{\operatornamewithlimits{NDCG}}
\newcommand*{\dcg}{\operatornamewithlimits{DCG}}
\newcommand*{\rel}{\operatornamewithlimits{relevance}}
\newcommand*{\ap}{\operatornamewithlimits{AP}}
\begin{document}

\title{Targeted display advertising: the case of preferential attachment}

% \IEEEpubid{0000--0000/00\$00.00˜\copyright˜2015 IEEE}
\author{\IEEEauthorblockN{Saurav Manchanda}
\IEEEauthorblockA{
\textit{University of Minnesota}\\
Twin Cities, USA \\
manch043@umn.edu}
\and
\IEEEauthorblockN{Pranjul Yadav}
\IEEEauthorblockA{
\textit{Criteo AI Lab}\\
Palo Alto, USA \\
p.yadav@criteo.com}
\and
\IEEEauthorblockN{Khoa Doan}
\IEEEauthorblockA{
\textit{Virginia Tech}\\
Arlington, USA \\
khoadoan@vt.edu}
\and
\IEEEauthorblockN{S. Sathiya Keerthi}
\IEEEauthorblockA{
\textit{Criteo AI Lab}\\
Palo Alto, USA \\
s.selvaraj@criteo.com}
}

\maketitle
% \IEEEpubidadjcol
\begin{abstract}
An average adult is exposed to hundreds of digital advertisements daily\footnote{https://www.mediadynamicsinc.com/uploads/files/PR092214-Note-only-150-Ads-2mk.pdf}, making the digital advertisement industry a classic example of a big-data-driven platform. As such, the ad-tech industry relies on historical engagement logs (clicks or purchases) to identify potentially interested users for the advertisement campaign of a partner (a seller who wants to target users for its products). The number of advertisements that are shown for a partner, and hence the historical campaign data available for a partner depends upon the budget constraints of the partner. Thus, enough data can be collected for the high-budget partners to make accurate predictions, while this is not the case with the low-budget partners. This skewed distribution of the data leads to \emph{preferential attachment} of the targeted display advertising platforms towards the high-budget partners. In this paper, we develop \emph{domain-adaptation} approaches to address the challenge of predicting interested users for the partners with insufficient data, i.e., the tail partners. Specifically, we develop simple yet effective approaches that leverage the similarity among the partners to transfer information from the partners with sufficient data to cold-start partners, i.e., partners without any campaign data. Our approaches readily adapt to the new campaign data by incremental fine-tuning, and hence work at varying points of a campaign, and not just the cold-start. We present an experimental analysis on the historical logs of a major display advertising platform\footnote{https://www.criteo.com/}. Specifically, we evaluate our approaches across $149$ partners, at varying points of their campaigns. Experimental results show that the proposed approaches outperform the other \emph{domain-adaptation} approaches at different time points of the campaigns. 
% Additionally, we also perform extensive analysis of the proposed approaches on the special case of partner cold-start, i.e., when no historical data is available for a partner, and show the advantage of the proposed approaches over the other competing approaches.
\end{abstract}

\begin{IEEEkeywords}
digital advertising, ad-click prediction, domain-adaptation, transfer-learning, cold-start

\end{IEEEkeywords}

\section{Introduction}
% \begin{figure}[h]
%   \centering
%   \includegraphics[width=0.8\linewidth]{ad_example.pdf}
%   \caption{The highlighted part shows an example of targeted display advertisement.}
%   \label{fig:advertisement}
% \end{figure}

The digital advertising industry aims to identify potentially interested users to show the product-related advertisements for a partner (a seller who wants to target users for its products). It has grown to be one of the most important forms of advertising as a consequence of the ubiquity of the internet and the increasing popularity of digital platforms. For example, nearly 170 billion U.S. dollars were spent on digital advertising in 2015 and this figure is projected to add up to more than 330 billion U.S dollars by 2021\footnote{https://www.statista.com/topics/1176/online-advertising/}. With such a growth rate, improving the advertisement experience for the partners and users is a valuable challenge for the digital advertising industry. 

The advertising platform pays to the publisher (the website on which the advertisement is displayed) for each advertisement it displays, but this investment is successful only if the user engages with the displayed advertisement. Thus, identifying the users who are most likely to \emph{engage} with a \emph{targeted advertisement} is fundamental to the digital advertisement industry. The digital advertising industry relies upon the historical engagement-logs to identify the users likely to engage with a given advertisement. Thus, it is of paramount importance to have sufficient and credible data to build accurate prediction models for user engagement prediction. 

\begin{figure}[t]
  \centering
  \includegraphics[width=0.9\linewidth]{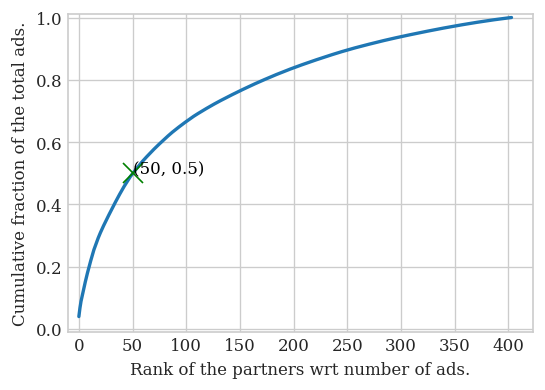}
  \caption{Advertisements distribution of the partners, for a random sample of $404$ partners, for a day. Less than one eighth (50 partners) of the partners drive $50\%$ of the advertisements.}
  \label{fig:distribution}
\end{figure}
The number of advertisements that can be shown for each partner, and hence the amount of training data available for each partner, depends upon its budget. This results in a skewed distribution of the data concerning the partners. Figure~\ref{fig:distribution} shows the distribution of advertisements in one day, for a random sample of $404$ partners. Less than one eighth (50 partners) of the partners drive $50\%$ of the advertisements. For the partners with sufficient budget, enough data can be collected to make accurate predictions, while this is not the case with the low-budget partners. As such, the digital advertisement industry suffers from \emph{preferential attachment} towards the high-budget partners, and hence, unfairness towards the low-budget partners. Although the individual budget of the low-budget partners is very small compared to that of the high-budget partners, the number of low-budget partners is considerably larger. Thus, the aggregated budget of these low-budget partners forms a considerable chunk of business opportunity. Hence, building approaches that are robust to \emph{preferential attachment} is an important challenge for the ad-tech industry, from both an ethical and business perspective.

To address this challenge, we developed \emph{domain-adaptation} approaches that leverage the similarity among the partners to transfer information from the head partners to the similar tail partners. In domain adaptation, we study two different (but related) domains, i.e., source and target. The domain adaptation task then consists of the transfer of knowledge from the source domain to the target domain. Specifically, our target domain is the non-campaign data, and the features in our target domain are the categories (such as electronics, apparel, etc.) in which a partner operates. Our source domain is the campaign data, in addition to the non-campaign data. The features corresponding to the campaign data are specifically engineered to the task of targeted-advertising using the domain knowledge, and these task-specific features are derived from upon the user-advertisement engagement counts from the campaigns (for example, how many times has the user engaged with the advertisement of a partner in last month). The prior domain adaptation approaches~\cite{ganin2016domain, motiian2017unified} focus on learning common representations that are discriminative as well as, invariant to the domains. However, among infinite such possible representations, the one that is closer to the source domain features should be preferred. This is because any machine learning algorithm depends upon the representation of the input data, and the source domain features are engineered for the task of targeted display advertising, thus are best-suited for the task. In this direction, we present two approaches that instead of learning the common representations, directly impute the source domain features using the target domain features. Our approaches assume that the partners with similar target domain representation should have similar source domain representation as well.  The two proposed approaches, \emph{Interpretable Anchored Domain Adaptation (IADA)} and \emph{Latent Anchored Domain Adaptation (LADA)} differ in the manner that IADA directly imputes the observed features in the source domain, while LADA imputes the features in the latent space, and hence is robust to the curse of dimensionality. 

We present an experimental analysis on the historical logs of a major display advertising platform\footnote{https://www.criteo.com/}. Specifically, we evaluate our approaches across $149$ partners, at different points of their campaign, i.e., we experiment with varying amounts of available data for the partners. Experimental results show that the proposed approaches outperform the baselines approaches at all points of the campaign, with LADA performing the best. Additionally, we also perform an extensive analysis of the proposed approaches on the special case of partner cold-start, i.e., when no historical data is available for a partner, and show the advantage of the proposed approaches over the competing approaches. For example, on the Mean Average Precision metric, LADA and IADA outperform the non-domain-adaptive baseline by $8.574\%$ and $2.710\%$ at cold-start, respectively.

\section{Related work}
The prior research that is most directly related to the work presented in this paper broadly spans the areas of domain-adaptation, transfer-learning, engagement prediction, and cold-start approaches. In this section, we review these areas.

\subsection{Domain Adaptation and Transfer Learning}\label{sec:da}
Domain adaptation and transfer learning are concerned with accounting for changes in data distributions between the training phase (source domain) and the test phase (target domain). Both domain adaptation and transfer learning have been used inconsistently and interchangeably within the machine learning literature. For this paper, we follow the definitions used in~\cite{arnold2007comparative}, which defines domain adaption as an instance of transfer learning when the prediction task across the domains is the same, but only the distribution of the data in the two domains is different. Transfer learning, on the other hand, refers to general-purpose knowledge transfer across domains and tasks. The domain adaptation methods can be broadly categorized into supervised, semi-supervised and unsupervised in consideration of labeled data of the target domain~\cite{wang2018deep}. %Supervised domain adaptation methods assume the availability of labeled data in the target domain. However, the labeled data are commonly not sufficient for the target domain task.  Semi-supervised domain adaptation methods assume the availability of both limited labeled data, and abundant unlabeled data in the target domain, which allows the methods to learn the structure information of the target domain. Unsupervised domain adaptation methods assume the unavailability of labeled data but have abundant unlabeled data available in the target domain. 
Domain adaptation has been made popular through its applications in computer vision~\cite{ganin2016domain, li2016deep, isola2017image, taigman2016unsupervised, shrivastava2017learning, bousmalis2017unsupervised, liu2017unsupervised, zhu2017unpaired} %, kim2017learning, yi2017dualgan, russo2018source, benaim2017one, wang2018high, murez2018image, huang2018multimodal, motiian2017few, motiian2017unified}
and natural language processing~\cite{ganin2016domain, ruder2019neural}. Domain adaptation and transfer learning have been applied in the field of engagement prediction as well. One of the earliest works~\cite{bickel2009transfer} derives a transfer learning procedure that produces resampling weights which match the pool of all examples to the target distribution of any given task.
\citeauthor{su2017improving}~\cite{su2017improving} improves the click prediction by transferring information from the data-rich product to a data-scarce target product.  \citeauthor{dalessandro2014scalable}~\cite{dalessandro2014scalable} uses the web browsing data of the users as the source domain to predict user's engagement. They use the prior learned from the source domain as a regularizer for the logistic regression on the target domain. \citeauthor{perlich2014machine}~\cite{perlich2014machine} uses the source domain to obtain task-aware representations of high-dimensional web browsing behavior and uses the learned representations to do predictions on the label-limited target domain. \citeauthor{aggarwal2019domain} uses the re-targeting platform as the source domain and uses the large amount of re-targeting data to cold-start the partners in the prospecting platform, which is their target domain. Although our work addresses the same challenge as~\cite{dalessandro2014scalable, perlich2014machine}, we go a step further and use the data from the frequent-advertised partners to improve the performance on the target domain. 
% In the recommender systems, there have been a few works using transfer learning which focus on imputing the missing features for new users in the target domain~\cite{oldridge2018adapting,he2018robust, pan2010transfer}, for diverse recommendations~\cite{pandey2018recommending}, or across domains~\cite{moreno2012talmud,azarbonyad2019domain}.

\subsection{Engagement Prediction}
Engagement prediction, such as \emph{Click-Through Rate (CTR)} prediction is the core challenge in targeted digital advertisement industry~\cite{mcmahan2013ad,richardson2007predicting}. 
% CTR is the number of clicks on a particular advertisement as compared to the number of impressions on it. It is an important measure to find the effectiveness of any online advertising campaign. 
Various linear and non-linear approaches have been proposed for CTR prediction. Popular approaches such as logistic regression~\cite{chapelle2015simple, mcmahan2013ad}, log-linear models~\cite{agarwal2010estimating} and decision trees~\cite{he2014practical} have shown decent performance in practice. Non-linear approaches to model the feature interactions include Factorization Machines (FMs) and deep learning approaches. Examples of factorization machines based approaches include~\cite{juan2016field, pan2018field, pan2016sparse}. Examples of deep learning approaches include the ones that model higher order interaction terms ~\cite{zhang2016deep,guo2017deepfm,liu2018field,cheng2016wide,zhou2018deep}, sequential models~\cite{zhou2018deep,ni2018perceive, zhang2014sequential}, and multimedia content based models~\cite{chen2016deep,zhai2016deepintent}. 
The multimedia-based models exploit newer sources of structured data like images and text in addition to traditional features. The other significant areas of research have been with using the keyword queries for optimization of the CTR in search and retrieval settings~\cite{edizel2017deep,regelson2006predicting,cheng2010personalized}. 

\subsection{Cold-start} Cold-start is the problem of being able to make predictions in the absence of data from the entity of interest. A prominent amount of work has been done in recommender systems and information retrieval to address the cold-start problem, examples of which include~\cite{sharma2019adaptive, wang2018cross, bobadilla2012collaborative, safoury2013exploiting, manchanda2019intent2, song2014transfer, manchanda2019intent}. As discussed in Section~\ref{sec:da}, some approaches~\cite{dalessandro2014scalable, perlich2014machine, aggarwal2019domain} have been proposed for cold-start problems in targeted digital advertising too. However, compared to these approaches, we go a step further and use the data from the frequent-advertised partners to improve the performance on the target domain.

\section{Problem Statement and Notation}

\begin{table}[!t]
\small
\centering
  \caption{Notation used throughout the paper.}
  \begin{tabularx}{\columnwidth}{lX}
    \hline
Symbol   & Description \\ \hline
$X^S$    & Input space in the source domain. \\
$X^t$    & Input space in the target domain. \\
$Y$    & Set of labels ($Y = $ \{Purchase, NoPurchase\}). \\
$y$    & A sample drawn from $Y$ ($y \in Y$). \\
$g$    & Domain transfer function to map target domain features in the discriminative new space. (can be target domain or domain-invariant space, depending upon the model) \\
$h$    & Domain transfer function to map source domain features in the latent space.\\
$f$    & Classification function that takes $g(x)$ as input and predicts if the user will engage with the advertisement. \\
$e$    & Classification function that takes $h(x)$ as input and predicts if the user will engage with the advertisement. \\
$\phi$    & Parameters of the function $g$ \\
$\theta$    & Parameters of the function $f$ \\
$\psi$    & Parameters of the function $h$ \\
$\upsilon$    & Parameters of the function $e$ \\
$\mathcal{L}_{M}^{a}$    & Loss function for estimating the function(s) $a$ for the model $M$. \\
$\alpha$    & Hyperparameter controlling the contribution of different loss functions in a multi-objective model. \\
\hline
\end{tabularx}
  \label{tab:notation}
\end{table}
We address the broad challenge of predicting if a user will purchase the product that is advertised to him/her or not. This is essentially a binary classification task, and we seek to estimate a mapping \emph{advertisement} $\rightarrow$ \{\emph{Purchase, No Purchase}\}. Each advertisement is part of some partner's advertisement campaign. The features for each advertisement are derived from the partners' prior campaign data and are specifically engineered to the task of targeted-advertising. Specifically, these engineered features are derived from the user-partner engagement counts (for example, the number of times the user has engaged with the partner in the last month). As such, there is not enough data to confidently estimate these engineered features for the tail partners. The specific case of cold-start corresponds to zero amount of data to estimate these features. Hence, estimating a classifier using these engineered features will be biased towards the head partners. 

We seek to improve the prediction performance for the tail partners, at different time points of their campaign. To achieve this, we present \emph{domain-adaptation} approaches that leverage the similarity among the partners to transfer information from the head partners to the similar tail partners. Formally, the domain adaptation task then consists of the transfer of knowledge from the source domain to the target domain. In our particular setting, the \emph{target domain} is the non-campaign data that correspond to the categories in which the partners operate (such as electronics, apparel, etc.). The \emph{source domain} consists of campaign data, in addition to the category data. As discussed earlier, the features for the campaign-data are engineered specific to the task and hence are best suited to predict the users' engagement. We denote the input space corresponding to the source domain as $\mathcal{X}^S$, and that to the target domain as $\mathcal{X}^T$. For simplicity, we assume that the $\mathcal{X}^S$ and $\mathcal{X}^T$ share the same feature space, but the probability distribution $P(\mathcal{X}^S) \neq P(\mathcal{X}^T)$. This simply corresponds to having zeros corresponding to the campaign data features in the target domain (since the only difference between the source and the target domain is the lack of campaign data features in the target domain). The approaches presented in this paper estimate a transformation function $g(\cdot)$ with parameters $\phi$ that takes as input a representation $x$ of an advertisement ($x$ can be sampled from either $\mathcal{X}^S$ or $\mathcal{X}^T$), and outputs an representation $g(x)$. The transformation function $g(\cdot)$ is responsible for performing the domain-adaptation, by encoding the source domain information in the target domain. The representation $g(x)$ is then given as an input to another function $f(\cdot)$ with parameters $\theta$, which outputs the probability $P(y|x)$, where $y\in $\{Purchase, No Purchase\}. 

The approaches presented in this paper first estimate a base model ($g(\cdot)$ and $f(\cdot)$) specifically to cold-start the partners. The base model is then incrementally updated as the campaign-data comes up. Thus, how well the fine-tuned model works during the later stages of a campaign depends upon how well we estimate the base model. Specifically, the base model is a neural-network, and we incrementally update it by fine-tuning with the new data. Given the amount of data generated by digital advertisement platforms, having models that can be easily fine-tuned is of utmost importance. 
Table \ref{tab:notation} provides a reference for the notation used throughout the paper.

\section{Background}
The prior work in the area of domain adaptation assumes that the predictors trained on the source domain are also good indicators on the target domain when the underlying distributions of the source and target domains are similar. Thus, these approaches focus on learning \emph{representations} ($g(x)$) that are \emph{discriminative} as well as, invariant to the domains. Specifically, these approaches minimize the following loss:
\begin{equation}
    \mathcal{L}_{DA}(\theta, \phi|x,y) = \mathcal{L}_{DA}^f(\theta, \phi|x,y) + \alpha\mathcal{L}_{DA}^g(\phi|x),
\end{equation}
where, $\mathcal{L}_{DA}^f$ is the \emph{discrimination} loss such as cross-entropy, and measures how well the estimated representation $g(x)$ is able to perform the classification in the source domain (in the target domain as well, as follows from the assumption of these approaches), $\mathcal{L}_{DA}^g$ is the loss with respect to the domain-invariance of the learned representation $g(x)$. Depending upon whether the target-domain labels are available or not, either \emph{unsupervised domain adaptation} or \emph{supervised domain adaptation} approaches can be used. Unsupervised domain adaptation approaches estimate domain-invariant representations irrespective of the target labels, while supervised domain adaptation approaches enforce domain invariance \emph{per-class}.
\subsection{Unsupervised Domain Adaptation}\label{sec:unsup}
Unsupervised domain adaptation approaches estimate domain-invariant representations irrespective of the target labels. The unsupervised approaches model $\mathcal{L}_{DA}^g$ as:
\begin{equation}
    \mathcal{L}_{DA}^g(\phi|x) = r(p_{x\sim \mathcal{X^S}}(g(x)), p_{x\sim \mathcal{X}^T}(g(x))),
\end{equation}
where $p_{x\sim \mathcal{X^S}}(g(x))$ is the probability distribution of $g(x)$, when $x$ is sampled from $\mathcal{X}^S$, $p_{x\sim\mathcal{X}^T}(g(x))$ is the probability distribution of $g(x)$, when $x$ is sampled from $\mathcal{X}^T$ and $r$ is a distance metric measuring how different are the two distributions. One of the popular approaches proposed for the unsupervised domain adaptation is \emph{Domain-Adversarial Neural Networks (DANN)}~\cite{ganin2016domain}. DANN estimates $\mathcal{L}^g_{DA}$ by maximising the discriminator loss of a binary classifier that separates the two domains using the representation $g(x)$. 

\subsection{Supervised Domain Adaptation}\label{sec:sup}
When the target domain labels are available, supervised domain adaptation (SDA) approaches~\cite{motiian2017unified} can estimate \emph{task-aware} representations. Specifically, as compared to the unsupervised approaches, these approaches try to estimate representations, which are domain-invariant \emph{per class}. Specifically, the \emph{domain-invariant} loss for supervised domain adaptation approaches can be written as
\begin{equation}
    \mathcal{L}_{DA}^g(\phi|x) = \sum_{i=1}^K (r(p_{x\sim \mathcal{X}^S| y=i}(g(x)), p_{x\sim \mathcal{X}^T| y=i}(g(x)))),
\end{equation}
where $p_{x\sim\mathcal{X}^S|y=i}(g(x))$ is the probability distribution of $g(x)$, when $x$ is sampled from $\mathcal{X}^S$ and the corresponding label of $x$ is $i$, $p_{x\sim\mathcal{X}^T|y=i}(g(x))$ is the probability distribution of $g(x)$, when $x$ is sampled from $\mathcal{X}^T$  and the corresponding label of $x$ is $i$; and $r$ is a distance metric measuring how different are the two distributions. %The SDA approaches such as~\cite{motiian2017unified} assume availability of scarce target domain data, which is insufficient to confidently minimize the classification error directly in the target domain

\section{Proposed approaches}

\begin{figure}[t]
  \centering
  \includegraphics[width=1.0\linewidth]{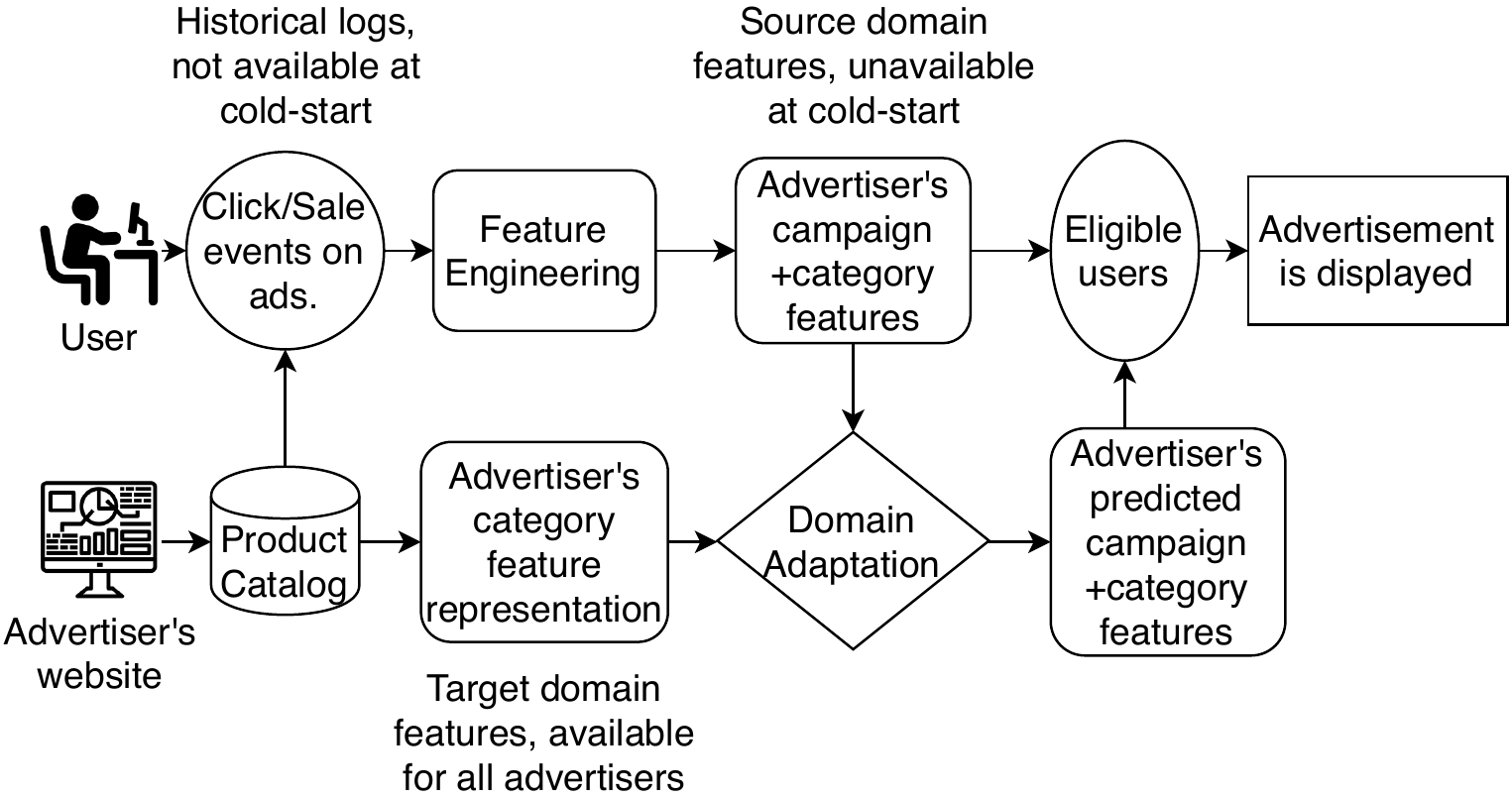}
  \caption{Work flow of data collection and proposed approaches.}
  \label{fig:workflow}
\end{figure}
The prior domain adaptation approaches~\cite{ganin2016domain, motiian2017unified} focus on learning common representations that are discriminative as well as, invariant to the domains. However, among infinite such possible representations, the one that is closer to the source domain features is preferred. This is because any machine learning algorithm depends upon the representation of the input data, and the source domain features are engineered for the task of targeted display advertising, thus are best-suited for the task at hand. One way to estimate a common representation that is closer to the source domain is to directly impute the source domain features using the target domain features. In other words, we assume that the partners with similar target domain representation have similar source domain representation as well. Since we have both the source and target domain features for the head partners, we learn the transformation function $g(\cdot)$ using the target domain features of head partners as input and source domain features as output. Once $g(\cdot)$ is learned, it can simply be applied on tail partners to predict their source domain representation. To this extent, we propose two approaches: \emph{Interpretable Anchored Domain Adaptation (IADA)} and \emph{Latent Anchored Domain Adaptation (LADA)}, that model the function $g(\cdot)$ to impute the features in the source domain, with the target domain features as input. The term \emph{anchored} refers to the manner, in which we aim to estimate the representations, i.e., they are \emph{anchored} to the source domain representation. The first approach, IADA, directly estimates the observed features in the source domain features, hence it estimates \emph{interpretable} representation in the source domain. The second approach, LADA, estimates a latent representation in the source domain, and hence is robust to the curse of dimensionality. Figure~\ref{fig:workflow} shows the workflow of the data collection process and how the proposed approaches apply to the gathered data. 

\subsection{Interpretable Anchored Domain Adaptation (IADA)}
IADA is a two-step algorithm: the first step performs the actual transfer of information and predicts the source domain features from the target domain features. The second step predicts whether the user engages with the given advertisement, using the predicted source domain features. Specifically, the first step learns the mapping function $g(\cdot)$ which transforms the target domain features to the source domain features. The second step later takes the output of the function $g(\cdot)$ as input and performs our core prediction task of whether the user will engage will the shown advertisement or not; i.e., the second step models the function $f(\cdot)$.

\subsubsection{Step 1} The loss function for the first step is given by 
\begin{equation}
    \mathcal{L}^g_{IADA}(\phi|x\sim\mathcal{X}^T, x\sim\mathcal{X}^S).
\end{equation}
Specifically, $\mathcal{L}^g_{IADA}$ is the loss with respect to predicting the source domain features $x\sim\mathcal{X}^S$ using only the target domain features $x\sim\mathcal{X}^T$. $\mathcal{L}^g_{IADA}$ can be a regression related loss, such as mean squared error (MSE). 

\subsubsection{Step 2} The loss function for the second step is given by 
\begin{equation}
    \mathcal{L}^f_{IADA}(\theta, \phi|g(x\sim\mathcal{X}^T), y),
\end{equation}
$\mathcal{L}^f_{IADA}$ can be any classification loss, such as cross entropy.

The functions can be jointly minimized as a linear combination of the two loss functions, i.e., 

\begin{multline}
    \mathcal{L}^{f,g}_{IADA} = \alpha\mathcal{L}^f_{IADA}(\theta, \phi|g(x\sim\mathcal{X}^T), y)
    \\+ (1-\alpha)\mathcal{L}^g_{IADA}(\phi|x\sim\mathcal{X}^T, x\sim\mathcal{X}^S),
\end{multline}
where $\alpha$ is a hyperparameter controlling the contribution of the individual loss components.

As we start getting the source domain data for fine-tuning, there is no need to model the $\mathcal{L}^g_{IADA}$ loss. Thus, to incrementally update the IADA model as we get the source domain data, only two modifications need to be done: (i) $x\sim\mathcal{X}^S$ is given as an input to $g(\cdot)$, the output of which is fed to $f(\cdot)$; and (ii) we only need to minimize the $\mathcal{L}^f_{IADA}(\theta, \phi|g(x\sim\mathcal{X}^T), y)$. In this paper, we implement both $g(\cdot)$ and $f(\cdot)$ functions as a multilayer perceptron, with two hidden layers.

\subsection{Latent Anchored Domain Adaptation (LADA)}
As compared to IADA, LADA imputes the source domain features, but in the latent space. Thus, training LADA involves one step in addition to IADA, i.e., estimating the latent representations of the head partners, to further act as supervision to learn the transformation function $g(\cdot)$. Particularly,  
LADA involves a three-step algorithm: the first step is to obtain a latent representation in the source domain for the head partners. The second step performs the information transfer and learns the mapping $g(\cdot)$ which maps the target domain features, to the latent source domain representation. The third step later takes the output of the function $g(\cdot)$ as input and performs our primary prediction task of whether the user will engage will the shown advertisement or not.

\subsubsection{Step 1} The objective here is to obtain a latent representation in the source domain for the partners for which we have data in the source domain, i.e., the head partners. To do so, we learn a mapping function $h(\cdot)$ which takes as input the source domain features (of the head partners) and gives as output a representation $h(x)$, where $x\sim\mathcal{X}^S$ is the input source domain features. The representation can be modeled as an output of the unsupervised approaches such as auto-encoders. However, to encode the task-specific information in the latent representation $h(x)$, we also leverage the labeled data of the head partners. In this direction, $h(x)$ is further passed onto a binary classifier $e(\cdot)$ which predicts whether the user will engage with the advertisement at hand or not. Specifically, the loss function corresponding to the first step is given by: 
\begin{equation}
    \mathcal{L}^{h,e}_{LADA}(\psi, \upsilon|x\sim\mathcal{X}^S, y),
\end{equation}
where $\psi$ are the parameters of the function $h$, and $\upsilon$ are the parameters of the function $e$. $\mathcal{L}^{h,e}_{LADA}$ can be any classification loss, such as cross-entropy. An example of how to model the first step is using a feed-forward network with at least one hidden layer, and any of the hidden layers corresponds to the latent representation we seek.

\subsubsection{Step 2} The loss function for the second step is given by 
\begin{equation}
    \mathcal{L}^g_{LADA}(\phi|x\sim\mathcal{X}^T, h(x\sim\mathcal{X}^S)),
\end{equation}
Specifically, $\mathcal{L}^g_{LADA}$ is the loss with respect to predicting the source domain features in the latent space ($h(x\sim\mathcal{X}^S)$) using only the target domain features $x\sim\mathcal{X}^T$; thus, constrains that the learned features $g(x\sim\mathcal{X}^T)$ are anchored to the latent source distribution. $\mathcal{L}^g_{LADA}$ can be a regression related loss, such as mean squared error (MSE). 

\subsubsection{Step 3} The loss function for the third step is given by 
\begin{equation}
    \mathcal{L}^f_{LADA}(\theta, \phi|g(x\sim\mathcal{X}^T), y),
\end{equation}
$\mathcal{L}^f_{LADA}$ can be any classification loss, such as cross entropy.

Step 1 if performed first to estimate the latent source domain representations of the head partners. Similar to IADA, the functions $f(\cdot)$ and $g(\cdot)$ can be jointly minimized as a linear combination of the two loss functions, i.e., 

\begin{multline}
    \mathcal{L}^{f,g}_{LADA} = \alpha\mathcal{L}^f_{LADA}(\theta, \phi|g(x\sim\mathcal{X}^T), y) 
    \\+ (1-\alpha) \mathcal{L}^g_{LADA}(\phi|x\sim\mathcal{X}^T, h(x\sim\mathcal{X}^S)),
\end{multline}
where $\alpha$ is a hyperparameter controlling the contribution of the individual loss components. 

Similar to the IADA model, to incrementally update the LADA model, only two modifications need to be done: (i) $x\sim\mathcal{X}^S$ is given as an input to $g(\cdot)$, the output of which is fed to $f(\cdot)$; and (ii) we only need to minimize the $\mathcal{L}^f_{LADA}(\theta, \phi|g(x\sim\mathcal{X}^T), y)$. The function $h(\cdot)$ plays no role for fine-tuning.

Like IADA, we implement both $g(\cdot)$ and $f(\cdot)$ functions as a multilayer perceptron, with two hidden layers. The functions $h(\cdot)$ and $e(\cdot)$ are jointly implemented as a single network. This network is a multilayer perceptron, with just one hidden layer. This hidden layer gives the latent representation $h(x)$ that we use for supervision in the later steps.

\section{Experimental methodology}\label{experiments}

\subsection{Dataset and Evaluation methodology}\label{sec:data}
We evaluate our methods on how well they can predict the users that are likely to engage with an advertisement of a partner, at all points during the campaign of that partner. We leverage the historical engagement logs of a major digital advertiser\footnote{https://www.criteo.com/} to estimate and evaluate our methods. As discussed before, our target domain is the non-campaign data, and the features in our target domain are the categories (such as electronics, apparel etc.) in which a partner operates. Our source domain is the campaign data, in addition to the non-campaign data. The features corresponding to the campaign data are specifically engineered to the task of targeted-advertising using domain knowledge. We generated our training, test and validation datasets using the historical engagement data as follows:
\begin{itemize}[leftmargin=*]
    \item We sampled $404$ partners and used their data to estimate and evaluate our methods. The data distribution for these partners is shown in Figure~\ref{fig:distribution}. We assume that the head-partner segment, i.e., the partners for which the majority of advertisements are displayed have reached the steady-state. Thus, from the $404$ sampled partners, we filtered the partners which correspond to $80\%$ of the total advertisements displayed, and these partners constituted our head-partners segment. A total of $218$ ($40\%$ of the total partners) constituted this segment. The remaining $186$ partners constituted our tail segment. From these  $186$ partners, we used data from the $37$ partners as the validation set to choose the hyperparameters and used data from the other $149$ partners to evaluate our methods. 
    \item We used a day of data from May 2019 for training (say, training day) and used the following day of data for evaluation (say, evaluation day). We first used all the data of the head partners from the training day to estimate our base models. Then, to incrementally update the base models, we randomly took $20\%$,  $40\%$,  $60\%$,  $80\%$ and $100\%$ of the data from each of the partners in the validation and test set, only from the training day, and used this data to fine-tune the base models.
\end{itemize}

\subsection{Performance Assessment metrics}\label{sec:metrics}
Engagement prediction is inherently a classification task. However, in the digital advertisement industry, because of the budget constraints, we are mainly interested in relatively few users that we consider most relevant for an advertisement. Thus, it makes sense to also consider engagement prediction as a ranking task. Thus, we consider classification, as well as ranking metrics, to evaluate our models. Specifically, we evaluate our approaches on the following three metrics:

\begin{itemize}[leftmargin=*]
    \item \textbf{Area Under Curve - Receiver Operating Characteristics} \\($\aucroc$): $\aucroc$ gives the probability that a randomly chosen positive example is deemed to have a higher probability of being positive than a randomly chosen negative example. It is one of the popular metrics for binary classification. 
    % Specifically, for a classifier $f:x\rightarrow y$, $\aucroc$ is defined as:
    % \begin{equation}
    %     \aucroc = \frac{1}{mn}\sum_{i=1}^{m}\sum_{i=1}^{n}\mathbbm{1}(f(x_i^+) > f(x_i^-)),
    % \end{equation}
    % where $m$ is the number of positive (engagement) examples, $n$ is the number of negative (non-engagement) examples and $\mathbbm{1}(\cdot)$ is the identity function. 
    
    \item \textbf{Normalized Discounted Cumulative Gain} ($\ndcg$)~\cite{wang2013theoretical}: $\ndcg$ is a popular measure of the ranking quality, that measures the gain (usefulness) of a prediction based on its position in the ranked result list. The gain is accumulated from the top of the result list to the bottom, with the gain of each result discounted at lower ranks. Instead of looking at all the results in the ranked list, usually $\ndcg$ is only calculated till a rank $k$, called as $\ndcg$@$k$. This makes sense as we are mainly interested in relatively few users that we consider most relevant for an advertisement, because of the budget constraints. In this paper, we report the results corresponding to $k=1,000$. Specifically, $\ndcg$ is obtained by normalizing Discounted Cumulative Gain ($\dcg$) of the ranking obtained as a result of the predictions made by the model which is supposed to be evaluated, with respect to the ideal ranking. $\dcg@k$ is given by
    \begin{equation}
        \dcg@k = \sum_{i=1}^{k}\frac{\rel_i}{\log_2(i+1)}
    \end{equation}
    where $\rel_i = 1$ if the prediction at $i$th rank is relevant, i.e., the predicted user at the $i$th rank engages with the user, and $\rel_i = 0$ otherwise. The logarithm in the denominator corresponds to discounting the gains at the lower ranks.
    
    \item \textbf{Average Precision} ($\ap$)~\cite{zhu2004recall}: $\ap$ is also a very popular performance measure in information retrieval. $\ap$ summarizes the precision-recall curve as a single number, by computing the average value of precision $p(r)$ as the recall $r$ changes from $r=0$ to $r=1$. This corresponds to the area under the precision-recall curve which is given by $
        \ap = \int_{0}^{1}p(r) dr
    $. 
    As done in practice, we replace this integral with a finite sum over every position in the ranked sequence of predictions, i.e., we calculate $\ap$ as
    \begin{equation}
        \ap = \sum _{i=1}^{n}p(i)\Delta r(i),
    \end{equation}
    where $i$ is the rank of a prediction in the ranked list, $n$ is the length of the ranked list (total number of predictions), $p(i)$ is the precision at cut-off $i$ in the list, and $\Delta r(i)$ is the change in recall from position $k-1$ to $k$.

\end{itemize}

We report the above metrics in both the micro and macro settings. In the macro setting, we calculate the metric independently for each partner and then takes the  average (hence treating all partners equally), whereas, in the micro setting, we aggregate the predictions for all the partners, and compute the metric on these combined predictions. Our primary setting of interest is the macro setting, as the micro setting can be biased towards the partners with a relatively larger volume of data.

\subsection{Baselines}
We use two different baselines to evaluate our approaches, as described below:
\begin{itemize}[leftmargin=*]
    \item \textbf{No transfer (NT)}: For the NT baseline, we directly make predictions using the target domain features, i.e., there is no transfer from the source domain. This baseline resembles the methods such as~\cite{dalessandro2014scalable, perlich2014machine}, which do not use the data from the frequent-advertised partners to improve the performance on the target domain. For a fair comparison, we implement NT as a multilayer perceptron, with two hidden layers; in the same manner as IADA and LADA.
    \item \textbf{Supervised Domain Adaptation (SDA)}: Traditional SDA approaches assume sparse unavailability of the labeled target domain data. As such, they do not explicitly model the classification loss in the target domain. This is not the case with us, because we have labeled data in the target domain as well. Thus, it would be unfair to directly compare the prior SDA approaches such as SDA-CCSA to IADA and LADA which also leverage the availability of the target domain data. Thus, we construct an SDA baseline that also leverages the target domain information. Specifically, our SDA baseline minimizes the following loss:
    \begin{multline}
        \mathcal{L}^{f,g}_{SDA} = \alpha(\mathcal{L}^f_{SDA}(\theta, \phi|g(x\sim\mathcal{X}^T), y)
        \\+ \mathcal{L}^f_{SDA}(\theta, \phi|g(x\sim\mathcal{X}^S), y)) 
        \\+  (1-\alpha) \mathcal{L}^g_{SDA}(\phi|x\sim\mathcal{X}^T, x\sim\mathcal{X}^S),
    \end{multline}
where $\alpha$ is a hyperparameter controlling the contribution of the individual loss components, and $\mathcal{L}^g_{SDA}(\phi|x\sim\mathcal{X}^T, x\sim\mathcal{X}^S)$ is implemented as the MSE loss between the representations $g(x\sim\mathcal{X}^T)$ and $g(x\sim\mathcal{X}^S)$ of the same advertisement.
\end{itemize}
\subsection{Parameter selection}

The train/test/validation splits were created as per the process mentioned in Section~\ref{sec:data}. For all the neural networks (IADA, LADA, and baselines), the number of nodes in the hidden dimension is set to $64$. For regularization, we used a dropout\cite{srivastava2014dropout} of $0.5$ between all layers, except between the penultimate and output layer. For optimization, we used the ADAM\cite{kingma2014adam} optimizer with the initial learning-rate set to 0.01. The tunable hyperparameter for all the approaches is $\lambda$, which is the contribution of various losses, depending upon the model. We tune $\lambda$ for all the approaches corresponding to the best performance on the macro setting. However, we independently choose $\lambda$ for each metric. The $\lambda$ was tuned using the grid search from the set \{0.1, 0.2, $\hdots$, 1.0\}. These chosen values for $\lambda$ are: (i) for IADA, $\lambda = 0.8$ for the $\aucroc$  and $\ap$ metrics, $\lambda = 0.5$ for the $\ndcg$ metric; (ii) for LADA, $\lambda = 0.9$ for the $\aucroc$  and $\ap$ metrics, $\lambda = 0.8$ for the $\ndcg$ metric; (iii) for SDA, $\lambda = 0.5$ for all the three metrics. To incrementally update the models, we use the optimizer from the same state, as it was when training of the base model finished.

\section{Results and Discussion}
We illustrate the performance of IADA and LADA and compare it with the other baselines in this section. First, we discuss the performance of various methods at different points of a campaign in Section~\ref{sec:journey}. Later we extensively analyze the particular case of cold-start, i.e., at the start of the campaign in Section~\ref{sec:cold_results}.
\subsection{Journey from the tail to head}\label{sec:journey}

\begin{figure*}[!t]
\centering
\subfloat[$\aucroc$ (macro)]{\includegraphics[width=0.33\textwidth]{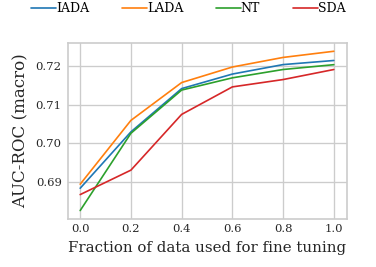}%
\label{fig:aucroc_macro}}
\hfil
\subfloat[$\ndcg$ (macro)]{\includegraphics[width=0.33\textwidth]{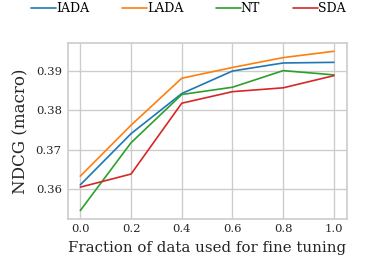}%
\label{fig:qual_2_seg_refine}}
\hfil
\subfloat[$\ap$ (macro)]{\includegraphics[width=0.33\textwidth]{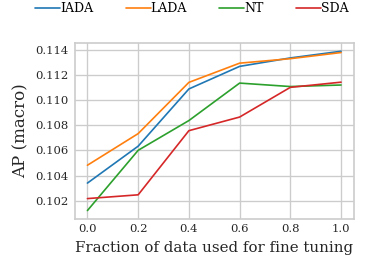}%
\label{fig:qual_2_mltm}}
\caption{Evaluation on the macro setting for different approaches. LADA and IADA constantly outperform the other baselines with LADA performing the best. The reported results are average over the $10$ runs, with different seed initialization.}
\label{fig:macro_metrics}
\end{figure*}

\begin{figure*}[!t]
\centering
\subfloat[$\aucroc$ (micro)]{\includegraphics[width=0.33\textwidth]{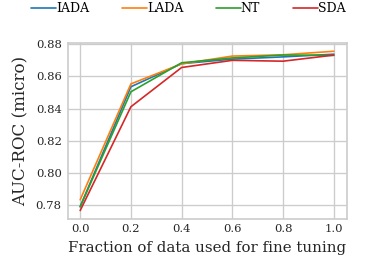}%
\label{fig:aucroc_macro}}
\hfil
\subfloat[$\ndcg$ (micro)]{\includegraphics[width=0.33\textwidth]{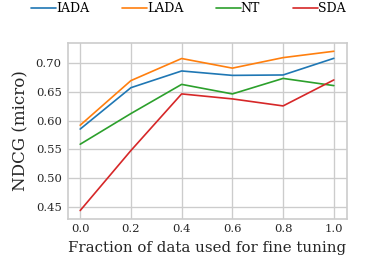}%
\label{fig:qual_2_seg_refine}}
\hfil
\subfloat[$\ap$ (micro)]{\includegraphics[width=0.33\textwidth]{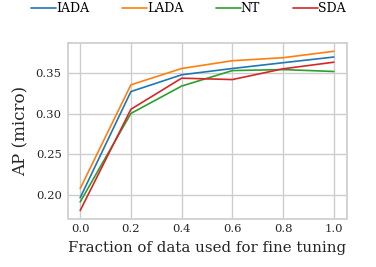}%
\label{fig:qual_2_mltm}}
\caption{Evaluation on the micro setting for different approaches. LADA and IADA constantly outperform the other baselines with LADA performing the best. The reported results are average over the $10$ runs, with different seed initialization.}
\label{fig:micro_metrics}
\end{figure*}
Figures~\ref{fig:macro_metrics} and~\ref{fig:micro_metrics} show the performance of all the approaches on the $\aucroc$, $\ndcg$ and $\ap$ metrics, for the macro and micro setting, respectively. The $x-$axis corresponds to the fraction of the data used to incrementally update the models, thus shows the different time points of a campaign. As discussed in Section~\ref{sec:data}, we incrementally update the models by fine-tuning them using a fraction of data of the validation and test partners, but only from the training day. For all the metrics, and both settings (macro and micro), the performance of all the methods generally gets better as the campaign goes on, i.e., as we get more campaign data to fine train the models. For all the metrics, and both settings (macro and micro), the proposed approaches LADA and IADA outperform all the baselines at all time points of a campaign. On the $\aucroc$ metric, the performance improvement of LADA over the NT baseline at the beginning of the campaign (cold-start) is $0.998\%$ and $0.563\%$, with respect to the macro and micro settings, respectively. As the campaign goes on, LADA keeps consistently outperforming the other baselines. When the complete data of a day is used to fine-tune the models ($x=1$ in Figures~\ref{fig:macro_metrics} and~\ref{fig:micro_metrics}), the performance improvement of LADA over the NT baseline on the $\aucroc$ metric is $0.482\%$ and $0.293\%$, for the macro and micro settings, respectively.

Similarly, the performance improvement of IADA on the $\aucroc$ metric over the NT baseline at the beginning of the campaign (cold-start) is $0.845\%$ and $0.019\%$, with respect to the macro and micro settings, respectively. As the campaign goes on, IADA keeps consistently outperforming the other baselines but LADA. When the complete data of a day is used to fine-tune the models, the performance improvement of IADA over the NT baseline on the $\aucroc$ metric is $0.150\%$ and $0.084\%$, with respect to the macro and micro settings, respectively.

On the $\ndcg$ and $\ap$ metrics, we see a similar trend as $\aucroc$, even in a more pronounced manner. On the $\ap$ metric, the performance improvement of LADA over the NT baseline at the beginning of the campaign is $3.576\%$ and $8.574\%$, with respect to the macro and micro settings, respectively. When the complete data of a day is used to fine-tune the models, the performance improvement of LADA over the NT baseline on the $\ap$ metric is $2.331\%$ and $7.072\%$, with respect to the macro and micro settings, respectively. Similarly, the performance improvement of IADA on the $\ap$ metric over the NT baseline at the beginning of the campaign is $2.167\%$ and $2.710\%$, with respect to the macro and micro settings, respectively. When the data of a day is used to fine-tune the models, the performance improvement of IADA over the NT baseline on the $\aucroc$ metric is $2.423\%$ and $5.057\%$, with respect to the macro and micro settings, respectively. 

Higher gain on the $\ap$ and $\ndcg$ metrics as compared to the $\aucroc$ metric is a result of the class-imbalance. In targeted-advertising, the ratio of advertisements that are engaged by a user, as compared to the total advertisements displayed is usually less than $5\%$. Consequently, a large change in the number of false positives can lead to a small change in the false positive rate used in ROC analysis; thus explaining small gain on the $\aucroc$ metric. On the other hand, Precision, and hence the $\ap$ metric, is robust to the class-imbalance problem~\cite{davis2006relationship}. 

We see a unique interesting pattern with the SDA approach. It usually performs better than the NT baseline in the macro setting for cold starting the partner, but at later points of the campaign, its performance, although increases with time, but the rate of increase is not at par with the NT baseline, thus NT performs better than the SDA as the campaign goes ahead in time. The reason for this lies in the design of the transformation function $g(\cdot)$. For SDA, the function $g(\cdot)$, maps both the source and target domain data into some common representation. For cold start, this leads to performance improvement as the common representation encodes information from the source domain. However, even though it is given access to source domain data for fine-tuning, $g(\cdot)$ tends to ignore the extra information since it is trained to deal with the source domain data, by mapping that to a representation, that is common to the target domain, thus ignoring some signals from the source domain. However, the NT baseline does not have this limitation. Although it performs worse at cold start, owing to lack of source domain information, it readily adapts to the source domain, thus, performing better than SDA with time. On the other hand, the proposed approaches IADA and LADA not only perform better at cold-start, but also adapt easily with the source-domain data, and hence, performs best at all points of the campaign. This is because, the function $g(\cdot)$ in IADA and LADA, unlike SDA, is not trained to ignore the extra source domain information, thus, is easy to fine-tune. %For example, if source domain data is given as n input to $g(\cdot)$ for IADA, $g(\cdot)$ has 

\subsection{The special case of cold-start}\label{sec:cold_results}
\begin{table}[!t]
\small
\centering
  \caption{Results on the macro metrics for the particular case of cold-start ($x=0$ in Figure~\ref{fig:macro_metrics}).}
  \begin{threeparttable}
     \begingroup
        \setlength{\tabcolsep}{3.5pt}
  \begin{tabularx}{\columnwidth}{Xrrrrrr}
    \hline
    Model    & \multicolumn{2}{c}{$\aucroc$}  & \multicolumn{2}{c}{$\ndcg$}  & \multicolumn{2}{c}{$\ap$}\\ 
             & (actual)  & (gain)  & (actual)  & (gain) & (actual) & (gain)\\ \hline
    IADA&$0.688$&$0.845\%$&$0.361$&$1.832\%$&$0.103$&$2.167\%$\\
    LADA&$0.689$&$0.998\%$&$0.363$&$2.454\%$&$0.105$&$3.576\%$\\
    NT&$0.683$&$0.000\%$&$0.355$&$0.000\%$&$0.101$&$0.000\%$\\
    % DANN&$0.674$&$-1.211\%$&$0.347$&$-2.101\%$&$0.097$&$-4.428\%$\\
    SDA&$0.687$&$0.604\%$&$0.361$&$1.669\%$&$0.102$&$0.943\%$\\\hline
  \end{tabularx}
 \endgroup
      \begin{tablenotes}
         \item[1] The gain is reported with respect to the NT baseline.
         \item[2] The reported results are average over the $10$ runs, with different seed initialization.
      \end{tablenotes}
      \end{threeparttable}
  \label{tab:cold_macro}
\end{table}

\begin{table}[!t]
\small
\centering
  \caption{Results on the micro metrics for the particular case of cold-start ($x=0$ in Figure~\ref{fig:micro_metrics}).}
  \begin{threeparttable}
     \begingroup
        \setlength{\tabcolsep}{2.5pt}
  \begin{tabularx}{\columnwidth}{lrrrrrr}
    \hline
    Model    & \multicolumn{2}{c}{$\aucroc$}  & \multicolumn{2}{c}{$\ndcg$}  & \multicolumn{2}{c}{$\ap$}\\ 
             & (actual)  & (gain)  & (actual)  & (gain) & (actual) & (gain)\\ \hline
    IADA&$0.779$&$0.019\%$&$0.585$&$4.714\%$&$0.197$&$2.710\%$\\
    LADA&$0.783$&$0.563\%$&$0.592$&$5.910\%$&$0.208$&$8.574\%$\\
    NT&$0.779$&$0.000\%$&$0.559$&$0.000\%$&$0.191$&$0.000\%$\\
    % DANN&$0.750$&$-3.676\%$&$0.540$&$-3.282\%$&$0.194$&$1.408\%$\\
    SDA&$0.777$&$-0.297\%$&$0.444$&$-20.617\%$&$0.181$&$-5.550\%$\\\hline
  \end{tabularx}
 \endgroup
      \begin{tablenotes}
         \item[1] The gain is reported with respect to the NT baseline.
      \end{tablenotes}
      \end{threeparttable}
  \label{tab:cold_micro}
\end{table}

\begin{table}[!t]
\small
\centering
  \caption{Precision@k for the particular case of cold-start ($x=0$ in Figure~\ref{fig:micro_metrics}).}
  \begin{threeparttable}
     \begingroup
        \setlength{\tabcolsep}{4.5pt}
  \begin{tabularx}{\columnwidth}{Xrrrrrrr}
    \hline
    Model/k    & 50  & 100  & 200  & 500  & 1000  & 1500  & 2000\\ \hline
    IADA&$0.154$&$0.137$&$0.117$&$0.092$&$0.074$&$0.064$&$0.057$\\
    LADA&$0.156$&$0.143$&$0.126$&$0.094$&$0.074$&$0.063$&$0.057$\\
    NT&$0.151$&$0.135$&$0.116$&$0.091$&$0.073$&$0.064$&$0.057$\\
    % DANN&$0.138$&$0.124$&$0.109$&$0.088$&$0.071$&$0.062$&$0.056$\\
    SDA&$0.151$&$0.136$&$0.118$&$0.092$&$0.074$&$0.064$&$0.057$\\\hline
  \end{tabularx}
 \endgroup
      \begin{tablenotes}
         \item[1] The hyperparameters are the ones that gave the best performance on the $\ap$ metric, for the results shown in ~\ref{fig:micro_metrics}.
      \end{tablenotes}
      \end{threeparttable}
  \label{tab:cold_patk}
\end{table}

\begin{figure}[!t]
\centering
\subfloat[ROC curve for the tail partners]{\includegraphics[width=0.50\columnwidth]{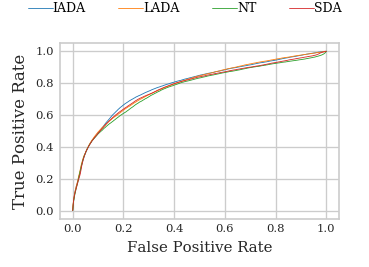}%
\label{fig:roc_all}}
\hfil
\subfloat[Truncated ROC curve for the tail partners]{\includegraphics[width=0.50\columnwidth]{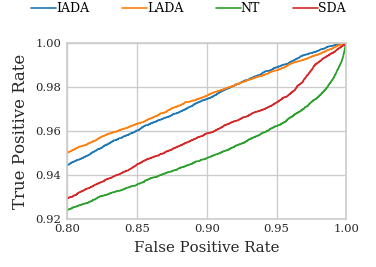}%
\label{fig:truncated_roc}}
\caption{ROC curves for the various methods evaluated on the tail partners.}
\label{fig:roc}
\end{figure}

Tables~\ref{tab:cold_macro} and~\ref{tab:cold_micro} show the performance of all the methods on all the metrics for the special case of the partner cold-start. As already discussed in Section~\ref{sec:journey}, the proposed approaches LADA and IADA outperform the other baselines. In this section, we take the discussion forward for cold-starting the partners. As outlined in Section~\ref{sec:metrics}, in the digital advertisement industry, we are mainly interested in a relatively few users that we consider most relevant for an advertisement because of the budget constraints. Hence, we also present results for the Precision@k metric. Precision@k represents the ratio of true positives in the predicted top-k positives. Table~\ref{tab:cold_patk} shows the Mean Precision@k metric, for different values of k and various methods. The metric is reported in the macro setting, i.e., it is reported as the average across all the tail partners. We see the same trend as the other metrics, with the proposed approaches LADA and IADA outperforming the other baselines, and LADA performing the best. As k increases, we observe that the Precision@k decreases for all the approaches. This can again be attributed to the class-imbalance problem. The number of true positives grows at a slower pace than the true negatives with an increase in k, thus the Precision@k tends to decrease as k increases. Besides, as k increases, the difference in the performance of different methods decreases. Moreover, as k reaches $1,000$, the Precision@k converges down to the same value for all the methods. The reason for this is: as k increases, the chances that the true positives of a partner are covered within the predicted top-k positives of any model increases. Thus, as k increases, partners tend to contribute equally to the Precision@k metric for all the models, leading to the same value of Precision@k for all the models if k is large. 

In addition, we also perform ROC analysis for the partner cold-start for all the approaches. Figure~\ref{fig:roc_all} shows the ROC plot, i.e., the plot of True Positive Rate with respect to the False Positive Rate. An ideal model would yield a point in the upper left corner or coordinate (0,1) of the ROC space. The plot-lines of all the methods lie pretty close to each other with IADA and LADA giving marginally better curves as compared to the baselines. As discussed earlier, the marginal advantage of IADA and LADA in the ROC space is a result of the class-imbalance problem. A large change in the number of false positives can lead to a small change in the false positive rate. However, because of the budget constraints, it can be more useful to focus on the region of the ROC space with a high true positive rate, i.e., few top users who are most likely to engage with an advertisement. Thus, we look at the truncated ROC space in Figure~\ref{fig:roc}, i.e., the top-right corner of the ROC space. The truncated ROC plot shows a very clear pattern, with IADA and LADA outperforming the other baselines, illustrating the advantage of IADA and LADA over the baselines.

\section{Conclusion and Future work}

In this paper, we address the challenge of predicting interested users for the tail partners in the digital advertising industry. Towards that, we developed two domain adaptation approaches that leverage the similarity among the partners to transfer information from the partners with sufficient data to similar partners with insufficient data. As compared to other domain-adaptation approaches, that estimate the common discriminative representations between the source and target domain, our proposed approaches directly impute the source domain features using the target domain features. The two proposed approaches, Interpretable Anchored Domain Adaptation (IADA) and Latent Anchored Domain Adaptation (LADA) differ in the manner that IADA directly imputes the observed features in the source domain, while LADA imputes the features in the latent domain, and hence is robust to the curse of dimensionality.

To the best of our knowledge, ours is the first attempt at using domain-adaptation approaches to transfer information from the head partners to the tail partners in the digital advertising industry. We envision that the proposed approaches will serve as a motivation for other applications that also suffer through preferential attachment. %As a future work, a promising direction is to explore Generative Adversarial Networks (GANs) to impute the source domain features.

% \section*{Acknowledgment}

% The preferred spelling of the word ``acknowledgment'' in America is without 
% an ``e'' after the ``g''. Avoid the stilted expression ``one of us (R. B. 
% G.) thanks $\ldots$''. Instead, try ``R. B. G. thanks$\ldots$''. Put sponsor 
% acknowledgments in the unnumbered footnote on the first page.

% \section*{References}

\bibliographystyle{IEEEtranN}  
{\footnotesize
\bibliography{refs}

% Generated by IEEEtranN.bst, version: 1.14 (2015/08/26)
\begin{thebibliography}{49}
\providecommand{\natexlab}[1]{#1}
\providecommand{\url}[1]{#1}
\csname url@samestyle\endcsname
\providecommand{\newblock}{\relax}
\providecommand{\bibinfo}[2]{#2}
\providecommand{\BIBentrySTDinterwordspacing}{\spaceskip=0pt\relax}
\providecommand{\BIBentryALTinterwordstretchfactor}{4}
\providecommand{\BIBentryALTinterwordspacing}{\spaceskip=\fontdimen2\font plus
\BIBentryALTinterwordstretchfactor\fontdimen3\font minus
  \fontdimen4\font\relax}
\providecommand{\BIBforeignlanguage}[2]{{%
\expandafter\ifx\csname l@#1\endcsname\relax
\typeout{** WARNING: IEEEtranN.bst: No hyphenation pattern has been}%
\typeout{** loaded for the language `#1'. Using the pattern for}%
\typeout{** the default language instead.}%
\else
\language=\csname l@#1\endcsname
\fi
#2}}
\providecommand{\BIBdecl}{\relax}
\BIBdecl

\bibitem[Ganin et~al.(2016)Ganin, Ustinova, Ajakan, Germain, Larochelle,
  Laviolette, Marchand, and Lempitsky]{ganin2016domain}
Y.~Ganin, E.~Ustinova, H.~Ajakan, P.~Germain, H.~Larochelle, F.~Laviolette,
  M.~Marchand, and V.~Lempitsky, ``Domain-adversarial training of neural
  networks,'' \emph{The Journal of Machine Learning Research}, vol.~17, no.~1,
  pp. 2096--2030, 2016.

\bibitem[Motiian et~al.(2017)Motiian, Piccirilli, Adjeroh, and
  Doretto]{motiian2017unified}
S.~Motiian, M.~Piccirilli, D.~A. Adjeroh, and G.~Doretto, ``Unified deep
  supervised domain adaptation and generalization,'' in \emph{Proceedings of
  the IEEE International Conference on Computer Vision}, 2017.

\bibitem[Arnold et~al.(2007)Arnold, Nallapati, and
  Cohen]{arnold2007comparative}
A.~Arnold, R.~Nallapati, and W.~W. Cohen, ``A comparative study of methods for
  transductive transfer learning.'' in \emph{ICDM Workshops}, 2007.

\bibitem[Wang and Deng(2018)]{wang2018deep}
M.~Wang and W.~Deng, ``Deep visual domain adaptation: A survey,''
  \emph{Neurocomputing}, vol. 312, pp. 135--153, 2018.

\bibitem[Li et~al.(2016)Li, Zuo, and Zhang]{li2016deep}
M.~Li, W.~Zuo, and D.~Zhang, ``Deep identity-aware transfer of facial
  attributes,'' \emph{arXiv preprint arXiv:1610.05586}, 2016.

\bibitem[Isola et~al.(2017)Isola, Zhu, Zhou, and Efros]{isola2017image}
P.~Isola, J.-Y. Zhu, T.~Zhou, and A.~A. Efros, ``Image-to-image translation
  with conditional adversarial networks,'' in \emph{Proceedings of the IEEE
  conference on computer vision and pattern recognition}, 2017.

\bibitem[Taigman et~al.(2016)Taigman, Polyak, and
  Wolf]{taigman2016unsupervised}
Y.~Taigman, A.~Polyak, and L.~Wolf, ``Unsupervised cross-domain image
  generation,'' \emph{arXiv preprint arXiv:1611.02200}, 2016.

\bibitem[Shrivastava et~al.(2017)Shrivastava, Pfister, Tuzel, Susskind, Wang,
  and Webb]{shrivastava2017learning}
A.~Shrivastava, T.~Pfister, O.~Tuzel, J.~Susskind, W.~Wang, and R.~Webb,
  ``Learning from simulated and unsupervised images through adversarial
  training,'' in \emph{Proceedings of the IEEE CVPR}, 2017, pp. 2107--2116.

\bibitem[Bousmalis et~al.(2017)Bousmalis, Silberman, Dohan, Erhan, and
  Krishnan]{bousmalis2017unsupervised}
K.~Bousmalis, N.~Silberman, D.~Dohan, D.~Erhan, and D.~Krishnan, ``Unsupervised
  pixel-level domain adaptation with generative adversarial networks,'' in
  \emph{Proceedings of the IEEE CVPR}, 2017, pp. 3722--3731.

\bibitem[Liu et~al.(2017)Liu, Breuel, and Kautz]{liu2017unsupervised}
M.-Y. Liu, T.~Breuel, and J.~Kautz, ``Unsupervised image-to-image translation
  networks,'' in \emph{NIPS}, 2017, pp. 700--708.

\bibitem[Zhu et~al.(2017)Zhu, Park, Isola, and Efros]{zhu2017unpaired}
J.-Y. Zhu, T.~Park, P.~Isola, and A.~A. Efros, ``Unpaired image-to-image
  translation using cycle-consistent adversarial networks,'' in
  \emph{Proceedings of the IEEE international conference on computer vision},
  2017.

\bibitem[Ruder(2019)]{ruder2019neural}
S.~Ruder, ``Neural transfer learning for natural language processing,'' Ph.D.
  dissertation, NATIONAL UNIVERSITY OF IRELAND, GALWAY, 2019.

\bibitem[Bickel et~al.(2009)Bickel, Sawade, and Scheffer]{bickel2009transfer}
S.~Bickel, C.~Sawade, and T.~Scheffer, ``Transfer learning by distribution
  matching for targeted advertising,'' in \emph{Advances in neural information
  processing systems}, 2009, pp. 145--152.

\bibitem[Su et~al.(2017)Su, Jin, Chen, Sun, Yang, Qiao, Xia, and
  Xu]{su2017improving}
Y.~Su, Z.~Jin, Y.~Chen, X.~Sun, Y.~Yang, F.~Qiao, F.~Xia, and W.~Xu,
  ``Improving click-through rate prediction accuracy in online advertising by
  transfer learning,'' in \emph{Proceedings of the International Conference on
  Web Intelligence}.\hskip 1em plus 0.5em minus 0.4em\relax ACM, 2017, pp.
  1018--1025.

\bibitem[Dalessandro et~al.(2014)Dalessandro, Chen, Raeder, Perlich,
  Han~Williams, and Provost]{dalessandro2014scalable}
B.~Dalessandro, D.~Chen, T.~Raeder, C.~Perlich, M.~Han~Williams, and
  F.~Provost, ``Scalable hands-free transfer learning for online advertising,''
  in \emph{Proceedings of the 20th ACM KDD}.\hskip 1em plus 0.5em minus
  0.4em\relax ACM, 2014, pp. 1573--1582.

\bibitem[Perlich et~al.(2014)Perlich, Dalessandro, Raeder, Stitelman, and
  Provost]{perlich2014machine}
C.~Perlich, B.~Dalessandro, T.~Raeder, O.~Stitelman, and F.~Provost, ``Machine
  learning for targeted display advertising: Transfer learning in action,''
  \emph{Machine learning}, vol.~95, no.~1, pp. 103--127, 2014.

\bibitem[Aggarwal et~al.(2019)Aggarwal, Yadav, and Keerthi]{aggarwal2019domain}
K.~Aggarwal, P.~Yadav, and S.~S. Keerthi, ``Domain adaptation in display
  advertising: an application for partner cold-start,'' in \emph{Proceedings of
  the 13th ACM Conference on Recommender Systems}, 2019, pp. 178--186.

\bibitem[McMahan et~al.(2013)McMahan, Holt, Sculley, Young, Ebner, Grady, Nie,
  Phillips, Davydov, Golovin, et~al.]{mcmahan2013ad}
H.~B. McMahan, G.~Holt, D.~Sculley, M.~Young, D.~Ebner, J.~Grady, L.~Nie,
  T.~Phillips, E.~Davydov, D.~Golovin \emph{et~al.}, ``Ad click prediction: a
  view from the trenches,'' in \emph{Proceedings of the 19th ACM KDD}, 2013.

\bibitem[Richardson et~al.(2007)Richardson, Dominowska, and
  Ragno]{richardson2007predicting}
M.~Richardson, E.~Dominowska, and R.~Ragno, ``Predicting clicks: estimating the
  click-through rate for new ads,'' in \emph{Proceedings of the 16th
  international conference on World Wide Web}.\hskip 1em plus 0.5em minus
  0.4em\relax ACM, 2007.

\bibitem[Chapelle et~al.(2015)Chapelle, Manavoglu, and
  Rosales]{chapelle2015simple}
O.~Chapelle, E.~Manavoglu, and R.~Rosales, ``Simple and scalable response
  prediction for display advertising,'' \emph{ACM Transactions on Intelligent
  Systems and Technology (TIST)}, vol.~5, no.~4, p.~61, 2015.

\bibitem[Agarwal et~al.(2010)Agarwal, Agrawal, Khanna, and
  Kota]{agarwal2010estimating}
D.~Agarwal, R.~Agrawal, R.~Khanna, and N.~Kota, ``Estimating rates of rare
  events with multiple hierarchies through scalable log-linear models,'' in
  \emph{Proceedings of the 16th ACM SIGKDD international conference on
  Knowledge discovery and data mining}.\hskip 1em plus 0.5em minus 0.4em\relax
  ACM, 2010, pp. 213--222.

\bibitem[He et~al.(2014)He, Pan, Jin, Xu, Liu, Xu, Shi, Atallah, Herbrich,
  Bowers, et~al.]{he2014practical}
X.~He, J.~Pan, O.~Jin, T.~Xu, B.~Liu, T.~Xu, Y.~Shi, A.~Atallah, R.~Herbrich,
  S.~Bowers \emph{et~al.}, ``Practical lessons from predicting clicks on ads at
  facebook,'' in \emph{Proceedings of the Eighth International Workshop on Data
  Mining for Online Advertising}.\hskip 1em plus 0.5em minus 0.4em\relax ACM,
  2014, pp. 1--9.

\bibitem[Juan et~al.(2016)Juan, Zhuang, Chin, and Lin]{juan2016field}
Y.~Juan, Y.~Zhuang, W.-S. Chin, and C.-J. Lin, ``Field-aware factorization
  machines for ctr prediction,'' in \emph{Proceedings of the 10th ACM
  Conference on Recommender Systems}.\hskip 1em plus 0.5em minus 0.4em\relax
  ACM, 2016, pp. 43--50.

\bibitem[Pan et~al.(2018)Pan, Xu, Ruiz, Zhao, Pan, Sun, and Lu]{pan2018field}
J.~Pan, J.~Xu, A.~L. Ruiz, W.~Zhao, S.~Pan, Y.~Sun, and Q.~Lu, ``Field-weighted
  factorization machines for click-through rate prediction in display
  advertising,'' in \emph{Proceedings of the 2018 World Wide Web Conference},
  2018, pp. 1349--1357.

\bibitem[Pan et~al.(2016)Pan, Chen, Liu, Xu, Ma, and Lin]{pan2016sparse}
Z.~Pan, E.~Chen, Q.~Liu, T.~Xu, H.~Ma, and H.~Lin, ``Sparse factorization
  machines for click-through rate prediction,'' in \emph{2016 IEEE 16th
  International Conference on Data Mining (ICDM)}, 2016.

\bibitem[Zhang et~al.(2016)Zhang, Du, and Wang]{zhang2016deep}
W.~Zhang, T.~Du, and J.~Wang, ``Deep learning over multi-field categorical
  data,'' in \emph{European conference on information retrieval}, 2016.

\bibitem[Guo et~al.(2017)Guo, Tang, Ye, Li, and He]{guo2017deepfm}
H.~Guo, R.~Tang, Y.~Ye, Z.~Li, and X.~He, ``Deepfm: a factorization-machine
  based neural network for ctr prediction,'' \emph{arXiv preprint
  arXiv:1703.04247}, 2017.

\bibitem[Liu et~al.(2018)Liu, Tang, Li, Yu, Guo, He, and Zhang]{liu2018field}
W.~Liu, R.~Tang, J.~Li, J.~Yu, H.~Guo, X.~He, and S.~Zhang, ``Field-aware
  probabilistic embedding neural network for ctr prediction,'' in
  \emph{Proceedings of the 12th ACM Conference on Recommender Systems}.\hskip
  1em plus 0.5em minus 0.4em\relax ACM, 2018, pp. 412--416.

\bibitem[Cheng et~al.(2016)Cheng, Koc, Harmsen, Shaked, Chandra, Aradhye,
  Anderson, Corrado, Chai, Ispir, et~al.]{cheng2016wide}
H.-T. Cheng, L.~Koc, J.~Harmsen, T.~Shaked, T.~Chandra, H.~Aradhye,
  G.~Anderson, G.~Corrado, W.~Chai, M.~Ispir \emph{et~al.}, ``Wide \& deep
  learning for recommender systems,'' in \emph{Proceedings of the 1st workshop
  on deep learning for recommender systems}.\hskip 1em plus 0.5em minus
  0.4em\relax ACM, 2016, pp. 7--10.

\bibitem[Zhou et~al.(2018)Zhou, Zhu, Song, Fan, Zhu, Ma, Yan, Jin, Li, and
  Gai]{zhou2018deep}
G.~Zhou, X.~Zhu, C.~Song, Y.~Fan, H.~Zhu, X.~Ma, Y.~Yan, J.~Jin, H.~Li, and
  K.~Gai, ``Deep interest network for click-through rate prediction,'' in
  \emph{Proceedings of the 24th ACM SIGKDD International Conference on
  Knowledge Discovery \& Data Mining}.\hskip 1em plus 0.5em minus 0.4em\relax
  ACM, 2018, pp. 1059--1068.

\bibitem[Ni et~al.(2018)Ni, Ou, Liu, Li, Ou, Zeng, and Si]{ni2018perceive}
Y.~Ni, D.~Ou, S.~Liu, X.~Li, W.~Ou, A.~Zeng, and L.~Si, ``Perceive your users
  in depth: Learning universal user representations from multiple e-commerce
  tasks,'' in \emph{Proceedings of the 24th ACM SIGKDD}.\hskip 1em plus 0.5em
  minus 0.4em\relax ACM, 2018, pp. 596--605.

\bibitem[Zhang et~al.(2014)Zhang, Dai, Xu, Feng, Wang, Bian, Wang, and
  Liu]{zhang2014sequential}
Y.~Zhang, H.~Dai, C.~Xu, J.~Feng, T.~Wang, J.~Bian, B.~Wang, and T.-Y. Liu,
  ``Sequential click prediction for sponsored search with recurrent neural
  networks,'' in \emph{Twenty-Eighth AAAI Conference on Artificial
  Intelligence}, 2014.

\bibitem[Chen et~al.(2016)Chen, Sun, Li, Lu, and Hua]{chen2016deep}
J.~Chen, B.~Sun, H.~Li, H.~Lu, and X.-S. Hua, ``Deep ctr prediction in display
  advertising,'' in \emph{Proceedings of the 24th ACM international conference
  on Multimedia}.\hskip 1em plus 0.5em minus 0.4em\relax ACM, 2016, pp.
  811--820.

\bibitem[Zhai et~al.(2016)Zhai, Chang, Zhang, and Zhang]{zhai2016deepintent}
S.~Zhai, K.-h. Chang, R.~Zhang, and Z.~M. Zhang, ``Deepintent: Learning
  attentions for online advertising with recurrent neural networks,'' in
  \emph{Proceedings of the 22nd ACM SIGKDD international conference on
  knowledge discovery and data mining}.\hskip 1em plus 0.5em minus 0.4em\relax
  ACM, 2016, pp. 1295--1304.

\bibitem[Edizel et~al.(2017)Edizel, Mantrach, and Bai]{edizel2017deep}
B.~Edizel, A.~Mantrach, and X.~Bai, ``Deep character-level click-through rate
  prediction for sponsored search,'' in \emph{Proceedings of the 40th
  International ACM SIGIR}.\hskip 1em plus 0.5em minus 0.4em\relax ACM, 2017,
  pp. 305--314.

\bibitem[Regelson and Fain(2006)]{regelson2006predicting}
M.~Regelson and D.~Fain, ``Predicting click-through rate using keyword
  clusters,'' in \emph{Proceedings of the Second Workshop on Sponsored Search
  Auctions}, vol. 9623, 2006, pp. 1--6.

\bibitem[Cheng and Cant{\'u}-Paz(2010)]{cheng2010personalized}
H.~Cheng and E.~Cant{\'u}-Paz, ``Personalized click prediction in sponsored
  search,'' in \emph{Proceedings of the third ACM international conference on
  Web search and data mining}.\hskip 1em plus 0.5em minus 0.4em\relax ACM,
  2010, pp. 351--360.

\bibitem[Sharma and Karypis(2019)]{sharma2019adaptive}
M.~Sharma and G.~Karypis, ``Adaptive matrix completion for the users and the
  items in tail,'' in \emph{The World Wide Web Conference}.\hskip 1em plus
  0.5em minus 0.4em\relax ACM, 2019.

\bibitem[Wang et~al.(2018)Wang, Peng, Wang, Philip, Fu, and
  Hong]{wang2018cross}
X.~Wang, Z.~Peng, S.~Wang, S.~Y. Philip, W.~Fu, and X.~Hong, ``Cross-domain
  recommendation for cold-start users via neighborhood based feature mapping,''
  in \emph{International Conference on Database Systems for Advanced
  Applications}.\hskip 1em plus 0.5em minus 0.4em\relax Springer, 2018, pp.
  158--165.

\bibitem[Bobadilla et~al.(2012)Bobadilla, Ortega, Hernando, and
  Bernal]{bobadilla2012collaborative}
J.~Bobadilla, F.~Ortega, A.~Hernando, and J.~Bernal, ``A collaborative
  filtering approach to mitigate the new user cold start problem,''
  \emph{Knowledge-Based Systems}, vol.~26, pp. 225--238, 2012.

\bibitem[Safoury and Salah(2013)]{safoury2013exploiting}
L.~Safoury and A.~Salah, ``Exploiting user demographic attributes for solving
  cold-start problem in recommender system,'' \emph{Lecture Notes on Software
  Engineering}, vol.~1, no.~3, pp. 303--307, 2013.

\bibitem[Manchanda et~al.(2019{\natexlab{a}})Manchanda, Sharma, and
  Karypis]{manchanda2019intent2}
S.~Manchanda, M.~Sharma, and G.~Karypis, ``Intent term weighting in e-commerce
  queries,'' in \emph{Proceedings of the 28th ACM International Conference on
  Information and Knowledge Management}.\hskip 1em plus 0.5em minus 0.4em\relax
  ACM, 2019, pp. 2345--2348.

\bibitem[Song et~al.(2014)Song, Wang, Chen, and Wang]{song2014transfer}
Y.~Song, H.~Wang, W.~Chen, and S.~Wang, ``Transfer understanding from head
  queries to tail queries,'' in \emph{Proceedings of the 23rd ACM International
  Conference on Conference on Information and Knowledge Management}.\hskip 1em
  plus 0.5em minus 0.4em\relax ACM, 2014, pp. 1299--1308.

\bibitem[Manchanda et~al.(2019{\natexlab{b}})Manchanda, Sharma, and
  Karypis]{manchanda2019intent}
S.~Manchanda, M.~Sharma, and G.~Karypis, ``Intent term selection and refinement
  in e-commerce queries,'' \emph{arXiv preprint arXiv:1908.08564}, 2019.

\bibitem[Wang et~al.(2013)Wang, Wang, Li, He, Chen, and
  Liu]{wang2013theoretical}
Y.~Wang, L.~Wang, Y.~Li, D.~He, W.~Chen, and T.-Y. Liu, ``A theoretical
  analysis of ndcg ranking measures,'' in \emph{Proceedings of the 26th annual
  conference on learning theory (COLT 2013)}, vol.~8, 2013, p.~6.

\bibitem[Zhu(2004)]{zhu2004recall}
M.~Zhu, ``Recall, precision and average precision,'' \emph{Department of
  Statistics and Actuarial Science, University of Waterloo, Waterloo}, vol.~2,
  p.~30, 2004.

\bibitem[Srivastava et~al.(2014)Srivastava, Hinton, Krizhevsky, Sutskever, and
  Salakhutdinov]{srivastava2014dropout}
N.~Srivastava, G.~Hinton, A.~Krizhevsky, I.~Sutskever, and R.~Salakhutdinov,
  ``Dropout: A simple way to prevent neural networks from overfitting,''
  \emph{The Journal of Machine Learning Research}, vol.~15, no.~1, 2014.

\bibitem[Kingma and Ba(2014)]{kingma2014adam}
D.~P. Kingma and J.~Ba, ``Adam: A method for stochastic optimization,''
  \emph{arXiv preprint arXiv:1412.6980}, 2014.

\bibitem[Davis and Goadrich(2006)]{davis2006relationship}
J.~Davis and M.~Goadrich, ``The relationship between precision-recall and roc
  curves,'' in \emph{Proceedings of the 23rd ICML}.\hskip 1em plus 0.5em minus
  0.4em\relax ACM, 2006.

\end{thebibliography}
}

\end{document}